\documentclass[10pt,twocolumn,letterpaper]{article}

\usepackage[pagenumbers]{cvpr} 

\usepackage{graphicx}
\usepackage{amsmath}
\usepackage{amssymb}
\usepackage{booktabs}

\DeclareMathOperator*{\argmax}{arg\,max} 

\usepackage[pagebackref,breaklinks,colorlinks]{hyperref}
\graphicspath{{images/}} 

\usepackage[capitalize]{cleveref}
\crefname{section}{Sec.}{Secs.}
\Crefname{section}{Section}{Sections}
\Crefname{table}{Table}{Tables}
\crefname{table}{Tab.}{Tabs.}

\def\cvprPaperID{*****} 
\def\confName{CVPR}
\def\confYear{2022}

\begin{document}

\title{Detecting Backdoor Poisoning Attacks \\on Deep Neural Networks\\ by Heatmap Clustering}

\author{Lukas Schulth\\
	University of Konstanz\\
	\\
	{\tt\small lukas.schulth@gmail.com}
	\and
	Christian Berghoff\\
	Federal Office for \\
	Information Security\\
	{\tt\small christian.berghoff@bsi.bund.de }
	\and
	Matthias Neu\\
	Federal Office for \\
	Information Security\\
	{\tt\small matthias.neu@bsi.bund.de}
}
\maketitle

\begin{abstract}
    Predicitions made by neural networks can be fraudulently altered by so-called poisoning attacks. A special case are backdoor poisoning attacks. We study suitable detection methods and introduce a new method called Heatmap Clustering. There, we apply a $k$-means clustering algorithm on heatmaps produced by the state-of-the-art explainable AI method Layer-wise relevance propagation. The goal is to separate poisoned from un-poisoned data in the dataset.
	We compare this method with a similar method, called Activation Clustering, which also uses $k$-means clustering but applies it on the activation of certain hidden layers of the neural network as input. We test the performance of both approaches for standard backdoor poisoning attacks, label-consistent poisoning attacks and label-consistent poisoning attacks with reduced amplitude stickers. We show that Heatmap Clustering consistently performs better than Activation Clustering. However, when considering label-consistent poisoning attacks, the latter method also yields good detection performance.
\end{abstract}

\section{Introduction}
According to \cite{deeplearning_weather} we find ourselves in the third wave of artificial intelligence.
Deep neural networks are used in numerous fields like production, advertising, communication, biometrics and the automotive domain. 

As the size of networks increases, so does their performance, but this comes at the cost of computational efficiency and work for curating high-quality datasets. As a consequence, ML researchers tend to make heavy use of training data from external sources, use pre-trained networks or even outsource training entirely to a third party. All of this introduces opportunities for manipulating the datasets. Such manipulations can introduce so-called backdoors in the network in what is called a poisoning attack \cite{biggio2012poisoning}. In backdoor poisoning attacks, the attacker poisons the dataset by manipulating data points with a special trigger pattern. In an image classification task, this trigger pattern is a visual pattern affixed on the image. After training on the poisoned dataset, the network falsely classifies samples containing the trigger as a target class chosen by the attacker. In the absence of the trigger, the data point is classified correctly, thus detecting backdoor poisoning attacks is challenging.\\

\noindent \textbf{Our Contribution.} We present a novel method for detecting backdoor poisoning attacks by analyzing the decision strategies used by the neural network on different data points. To explore these strategies, we use Layer-wise Relevance Propagation (LRP) \cite{LRP_first_paper}, feed the resulting data representation into a clustering algorithm and use the Gromov-Wasserstein (GW) \cite{gromov2007metric} distance as distance measure. We demonstrate the performance of our approach and compare it to prior backdoor detection methods.

\section{Related Work}

\noindent \textbf{Poisoning Attacks.}
In standard backdoor attacks, the image label of poisoned data points is not consistent with its content as perceived by a human observer. Poisoned data points can in principle be detected by human inspection, but due to the large size of datasets a manual approach is infeasible. Label-consistent backdoor poisoning attacks \cite{labelconsistent} eliminate this inconsistency while still aiming for the network learning the backdoor trigger. As a result, these label-consistent backdoor poisoning attacks are even harder to detect. The authors also introduce methods to reduce the visibility of triggers by the human eye.  So far, there is no known detection method for label-consistent backdoor poisoning attacks to the best of our knowledge.\\

\noindent \textbf{Detection Methods.}
Various methods have been developed to detect poisoning attacks on neural networks. Most of these methods are motivated by the fact that poisoned data points are in some sense outliers and the neural networks must learn special strategies to process them correctly. The detection methods thus aim to use clustering methods to separate these special strategies from the inference used on unpoisoned data points. To this end, the methods leverage additional information.\\
Activation Clustering \cite{AC} uses PCA to reduce the dimension of activations of intermediate network layers before feeding them as an input to a clustering algorithm.
In \cite{spectral_signatures}, Spectral Signatures (SpecSig) are introduced. They are a property of all known backdoors and allow using tools from robust statistics to thwart the attacks. Specifically, the authors show that standard backdoor poisoning attacks tend to leave behind a detectable trace in the feature representation learned by the neural network. 

In \cite{imagenet_unhansed_v2} the detection of clever hans (CH) and backdoor artifacts using SpecSig is considered. Clever hans artifacts are \textit{spurious correlations} in the training data that are exploited by the model. The authors' experimental results show that SpecSig works well for detecting standard backdoor attacks. In the case of CH artifacts, however, SpecSig only produces results that are only slightly better than random guessing. Since the fundamental concept of label-consistent poisoning attacks and CH artifacts is the same (although the setting in poisoning attacks is adversarial, whereas CH artifacts are not introduced on purpose), we pursue the general approach from \cite{imagenet_unhansed_v2} and aim to detect all types of poisoning attacks by applying a $k$-means clustering method on data representations which are produced by Explainable Artificial Intelligence (XAI) methods. 
The idea of using an XAI method is inspired by the work of \cite{imagenet_unhansed_v1}, where CH artifacts are detected by inspecting which strategies the networks use to arrive at the same classification output.\\

\noindent \textbf{Explainable Artificial Intelligence.}
Explainable Artificial Intelligence deals with methods to explain models or single decisions of a model. 

Due to the complexity of neural networks, it is difficult to explain their inner functionality in a comprehensible manner. In discussing this issue, \cite{lipton} introduced the concepts of transparency (also called interpretability) and explainabilty. If a model is transparent, it is self-explanatory and one understands the model inputs.
If a model is too complex, the state of transparency cannot be reached. In that case, the understanding of the model can still be improved by using local or global explanation methods \cite{kistudie}.
Local methods aim at explaining the model behavior regarding single inputs. Important local methods include attribution-based methods (sensitivity analysis \cite{baehrens2010explain}, LRP \cite{LRP_first_paper}, DeepLIFT \cite{deeplift}, Integrated Gradients \cite{integratedgradients}, Grad-CAM \cite{gradcam}, Guided Backpropagation), SHAP values \cite{SHAP} and surrogate-based methods such as LIME \cite{lime}.
On the other hand, global methods try to improve the understanding of the model behavior as a whole. Global methods include Spectral Relevance Analysis (SpRAy) \cite{unmaskingCH}, Feature Visualization \cite{feature_vis} and Network Dissection \cite{networkdis}.

\section{Poisoning Attacks}
For a classification task performed by a neural network $f = f_\theta :\mathbb{R}^d \to \mathbb{R}^K$, let $d$ be the input dimension and $K$ the number of different classes. $\theta = (w,b)$ denotes the collection of parameters of the neural network.\\
For an input $x$ the predicted output label $y_{pred}$ is given by

\begin{equation}y_{pred} = \argmax_{j=1,...,K}{f_j(x)}\end{equation}
The network is trained on data pairs of images and labels $\mathcal{D} = \lbrace (x_i,y_i) | i=1,\ldots,N \rbrace$, where $N$ is the total number of images in the training dataset.

Attacks that are performed on fully trained networks during the inference phase in order to manipulate the model output are known as adversarial attacks and are extensively studied in the literature \cite{papernot2016transferability, kurakin2018adversarial, goodfellow2014explaining, evtimov2017robust}. In the setting of poisoning attacks \cite{biggio2012poisoning, shafahi2018poison, labelconsistent, badnets}, which is our concern in this article, the attack happens before the training process. If the attacker has access to the dataset, he can either manipulate existing data points or add additional ones.

In general, poisoning attacks aim to weaken the performance of the trained network either in general or just on a specific class.  Backdoor poisoning attacks introduce a backdoor trigger in a data point.  If the trigger is present in a data point at inference time, a network trained on the manipulated dataset will classify that data point as an element of the target class chosen by the attacker. If there is no backdoor trigger in the data point, it is classified as belonging to the original class.\\

Poisoning attacks can be executed on different data types. This paper focuses on poisoning attacks on image classification tasks.

\subsection{Label-consistent Poisoning Attacks}
In contrast to standard backdoor attacks, the image label for label-consistent attacks as introduced in \cite{labelconsistent} is not changed during an attack. In a label-consistent attack, the attacker chooses a target class and introduces a trigger in images of this class with the goal that the network learns the trigger instead of other information in these images. To achieve this behavior the attacker uses a second neural network as a surrogate model to deteriorate the data quality of the manipulated images before introducing the trigger. This step is necessary to get a successful attack with poisoned images which are still looking like the original images.
If the attack is successful, the trigger is learned by the neural network and can be applied to any image presented during inference time to force a wrong classification towards the target class.

\begin{figure}[ht]
	\centering
	\includegraphics[width=0.15\textwidth]{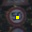}
	\includegraphics[width=0.15\textwidth]{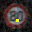}
	\caption{Examples of poisoned data points.}
	
	\label{im:example_pisoned_data}
\end{figure}

The poisoned samples $x_{adv}$ according to \cite{labelconsistent} are created by using a projective gradient descent method \cite{madry2017towards} as follows: 
\begin{equation}	
x_{adv} = \argmax_{||x'-x||_p \leq \varepsilon}{\mathcal{L}(x',y,\theta_2)}, \qquad 1 \leq p \leq \infty
\end{equation}
where $x$ is the original data point with label $y$, $\theta_2$ represents the parameters of the second neural network used by the attacker and $\mathcal{L}$ is the loss function.
The attacker thus creates a new data point whose distance to the original data point is bounded by $\varepsilon >0$ in the $p$-norm. As the last step, the trigger is placed on this new data point as mentioned above. Examples of poisoned data points from our experiments are shown in \autoref{im:example_pisoned_data}.

\subsection{Attacks with reduced amplitude stickers}
\begin{figure}[ht]
	\centering
	\includegraphics[width=0.15\textwidth]{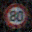}
	\includegraphics[width=0.15\textwidth]{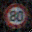}
	\includegraphics[width=0.15\textwidth]{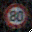}		
	\caption{Poisoned data points with reduced amplitude trigger in the lower right corner with amplitudes $amp=32,64,255$.}
	
	\label{im:LCPA}
\end{figure}

In order to make these label-consistent attacks even harder to detect, the use of so-called amplitude stickers has been proposed \cite{AC} to make the trigger less visible. Instead of changing the values of a number of pixels completely, a fixed amplitude value $amp \in \{0, \ldots, 255\}$ is added to all three channels of a fixed number of pixels. Reducing the amplitude value results in backdoor poisoning attacks which are even harder to detect and can barely be perceived by the human eye.
Examples of such amplitude stickers from our experiments are shown in \autoref{im:LCPA}.

\section{Defenses}
The goal of poisoning detection methods is to find the manipulated data samples to remove them from the dataset. In the following, we assume that we have one attacked class, which is poisoned by adding a small number of data points including a backdoor trigger, which results in a false classification of data points including this specific trigger.
Additionally we assume that the defender has already established which class is suspicious in the sense that it could possibly contain manipulated data points. Methods to find suspicious classes are discussed in \cite{imagenet_unhansed_v1}. 

Therefore, our main idea is to cluster the attacked class into two subsets: one containing the original data points and the other one containing the manipulated data points. To this end, an unsupervised clustering algorithm is used.\\

\noindent \textbf{Clustering methods}\\
The important characteristics of the clustering method are the data representation as well as the distance measure used, which is also used to compute the mean value for the $k$-means clustering.
We compare three different approaches:\\

\noindent \textbf{Clustering on raw image data}
The first approach applies $k$-means clustering directly on the image data. Each image is PCA-reduced before applying the $k$-means-algorithm.\\

\noindent \textbf{Activation Clustering}
Activation Clustering \cite{AC} uses the activations of one of the hidden layers of the neural network as the input of the clustering algorithm. This way, the clustering algorithm can take additional information which is more specific to the network under attack and the decision strategies it has learned into account.\\

\noindent \textbf{Heatmap Clustering}
Our novel approach extends Activation Clustering by using information that is more pertinent to the decision strategies learned by the neural network. More specifically, we use heatmaps generated by LRP as the image representation. We also modify the distance measure to consider both mass (relevances) and distance between points on the heatmaps for the definition of the distance between images. This defines the so-called Gromov-Wasserstein distance \cite{gromov2007metric, memoli2011gromov}. \\
In the following section, we explain both the calculation of heatmaps using the Layer-Wise Relevance Propagation method \cite{LRP_first_paper} and the modification of the distance measure. 

\section{Methods and Metrics}
In this section, we present metrics for evaluating poisoning attacks and countermeasures, the clustering algorithm, methods to create specific data representations as well as the dataset and neural network we used.

\subsection{Metrics for evaluating poisoning attacks}
In this section, we introduce a measure to express how successful a poisoning attack is. For a specific class of images we place the trigger in every image of the test dataset and count how many poisoned data points are classified correctly versus how many are assigned to the target class, as desired by the attacker. We refer to the rate of poisoned images classified as the target class as the \textit{attack success rate} (ASR).

To evaluate label-consistent poisoning attacks we introduce the selected \textit{trigger} to all images of all classes except the target class in the test dataset. We again compute the number of poisoned images per class that are classified as the target class. When compared to the total number of images, this defines a per-class ASR. We also define the \textit{mean attack success rate} (MASR), which considers all classes except the target class.

Following the procedure in \cite{labelconsistent}, we use trigger stickers with the full amplitude in the evaluation.

\subsection{Metrics for evaluating detection algorithms}
In order to quantify how well a defence method can detect the poisoning attacks, we again count the number of true positives (TP), true negatives (TN), false positives (FP) and false negatives (FN) from the binary classification results of the defence method.
We call the accuracy (ACC), true positive rate (TPR) and true negative rate (TNR) the detection rates.

\subsection{$k$-means Clustering}
In order to separate manipulated data from non-manipulated data in a suspicious class, we use $k$-means++ \cite{kmeans++}, 
which is a form of the $k$-means algorithm with better initialization of the first $k$ cluster centers. After picking the first cluster center, the $k-1$ remaining ones are not chosen completely at random, but depending on the maximum distance to the first one. This is intended to increase the convergence of the $k$-means algorithm.

\subsection{Layer-wise Relevance Propagation}
The concept of Layer-wise Relevance Propagation (LRP) allows computing a so-called heatmap. LRP was first introduced in \cite{LRP_first_paper} and later formally derived for deep neural networks\cite{LRP_DNN}.

The idea of LRP is to create a relationship between the output $f(x)$ of a neural network $f: \mathbb{R}^d \to \mathbb{R}^K$ and the input $x$. The output of the last layer of the neural network is interpreted as the relevance towards a specific class.
These relevance values are then propagated backwards from layer to layer through the network towards the input layer.

The algorithm preserves the total relevance towards a class $j$ while it is propagated from layer $l+1$ to layer $l$ in the following way:

\begin{equation*}
f_j(x) = ... = \sum_{d=1}^{V(l+1)}{R_d}^{(l+1)} = \sum_{d=1}^{V(l)}{R_d}^{(l)} = ... = \sum_{d=1}^{V(1)}{R_d^{(1)}}\label{erhaltungseigenschaft},
\end{equation*}
where $V(l)$ is the dimension of layer $l$ and $R_d^{(l)}$ are the relevances of the neurons in layer $l$. 
Various relevance propagation rules for different layers with different characteristics have been proposed \cite{lapuschkin, LRP_DNN}.

\begin{figure}[ht]
	\centering
	
	\includegraphics[scale=0.35]{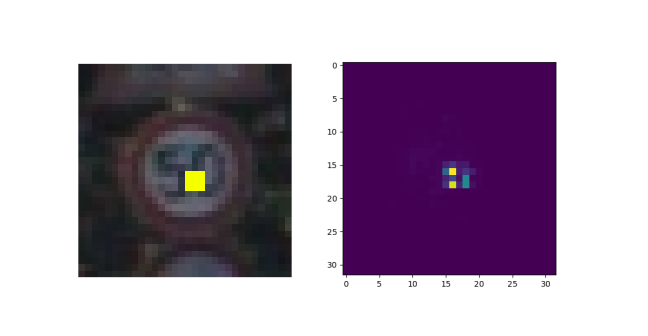}

	\caption{\textbf{Left:} Poisoned data point of the class 'speed limit 50 km/h'. \textbf{Right:} LRP output (relevances towards the class 'speed limit 80 km/h'.) }
	
	\label{im:example_poisoned_data_lrp}
\end{figure}

The result yielded by LRP is a relevance score for each input neuron towards a specific classification result.
The relevance scores of the input layer can easily be visualized as a heatmap by summing over all color channels of an image. The result for one poisoned image is shown in \autoref{im:example_poisoned_data_lrp}.

\noindent As a preprocessing step before applying clustering, we restrict each heatmap to the pixels accounting for the top $99\%$ of its relevance.

\begin{figure}[ht]
	\centering
	\includegraphics[width=1.0\linewidth]{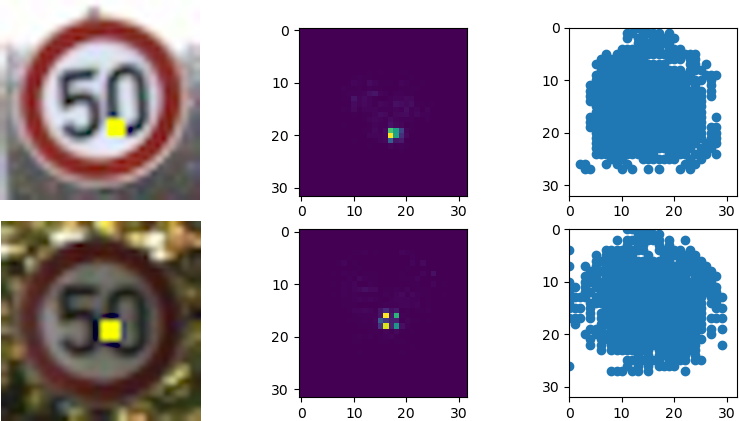}
	\caption{Two examples of poisoned input images, the resulting LRP heatmaps and point clouds.
		\textbf{Left}: Two input images of the class speed limit 50 sign with a green-yellow sticker of $3 \times 3$ pixels. \textbf{Middle}: LRP heatmap of speed limit 50 sign manipulated towards speed limit 80 class.  \textbf{Right}: Distribution of the most important pixel points, which make up $99 \%$ of the total relevance.}	\label{im:punktwolke}
\end{figure}

The heatmaps and the position of the remaining most relevant pixels are shown in \autoref{im:punktwolke} for two poisoned images.

\subsection{Gromov-Wasserstein-Distance}
In contrast to clustering on raw image data and Activation Clustering, for our clustering approach based on heatmaps we use the Gromov-Wasserstein distance (GWD) \cite{gromov2007metric, gwd_averaging_kernels, computationalOT}, which is a distance measure between metric-measures-spaces $(X,d_X,\mu_X)$, where $d_X$ is a distance measure and $\mu_X$ is a measure on the ground space $X$. The idea is to use $l^2$ as the distance and the relevance scores of a heatmap, calculated by LRP, define a measure on the ground space. We then want to compute distances between representations $(C_1,p_1)$ and $(C_2,p_2)$, where $C \in \mathbb{R}^{n_1 \times n_1}$ is the distance matrix of pair-wise $l^2$-distances between pixels and $p \in \mathbb{R}^{n_1}$ is the distribution of relevance over all pixels of an image.

Using the GWD we get a rotation- and translation-invariant distance measure \cite{vayer2020contribution}. This way we get a better distance measure in comparison to the euclidean distance in the sense, that images containing a backdoor trigger are still "close" even if the backdoor trigger is placed at different positions in two images.

While running the $k$-means algorithm we have to compute distances to all $k$ different means and recalculate $k$ different means in each iteration. In the case of the GWD the mean is a so-called Gromov-Wasserstein barycenter \cite{bary_wasserstein_space}.

\begin{table*}
	\centering
		\begin{tabular}{|c|c|ccc|ccc|}
			\hline
			Percentage of  	& ASR  		& 		&	Activation Clustering 	& 		& Proposed Method:	& Heatmap Clustering	& 		  \\
			poisoned data 		& 			& 		&	Detection Rate 	& 		& 			& Detection Rate	&		\\
			& 			& ACC 	& 	TPR 	& TNR  	& ACC 	& TPR 	& TNR 	 	\\\hline
			
			33			& 100	& 94.76 & 90.77 & 96.73 & \textbf{99.96}	& 99.88	& 100.00 \\ 
			15			& 100	& 76.51 & 99.31 &72.48 & \textbf{100.00}	& 100.00 	& 100.00\\
			10			& 100 	& 87.62 & 96.17 & 86.17& \textbf{99.29 }	& 92.9 	& 100.00 \\
			5			& 100	& 57.92 & 20.69 & 59.88 & \textbf{99.48 }	& 90.8		& 99.94 \\
			2			& 100	& \textbf{52.91} & 0.0 & 54.0 & 52.85		& 0.0 	& 53.94\\ \hline	 				
	\end{tabular}
	\caption[Comparison of attacks and defenses for SPAs]{Comparison of the performance of the detection of different attacks using Activation Clustering and Heatmap Clustering with $s=3$.}
	\label{tab:SPA_def_inv3_gwclustering}	
\end{table*}
\subsection{Network}
For our experiments we use one of the popular network architectures in image recognition called Inception \cite{goingdeeperwithconvolutions}. 
We use a modified version of the Inception-v3 architecture, which contains only one of the Inception modules. As a consequence, our network has only about one million trainable parameters as opposed to 24 million for the standard Inception-v3 network.\\

\noindent \textbf{Training.} For training, we used cross-entropy as the loss function and Adam \cite{adam} as the optimizer. We use a maximum number of 100 training epochs with early stopping that aborts the training process if the validation loss has not decreased any further for 20 epochs.\\

\noindent \textbf{Data augmentation.}
While training we use the following data augmentation to read the image data: RandomResizedCrop, RandomRotation, ColorJitter, RandomAffine, RandomGrayscale, Normalisation from the PyTorch toolbox \footnote{\url{https://pytorch.org/vision/main/transforms.html}}.

\subsection{Dataset}
\begin{figure}
	\centering
	\includegraphics[width=0.47\textwidth]{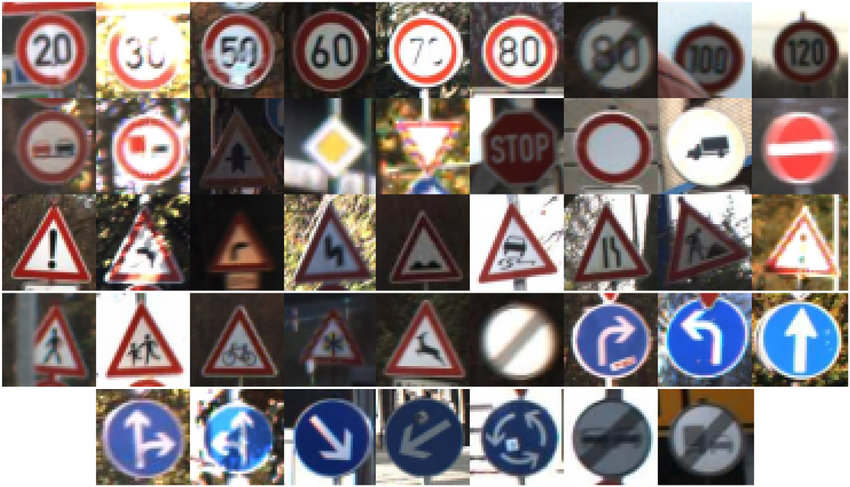}
	
	\caption{Images of all different classes in the German Traffic Sign Recognition Benchmark dataset.}
	\label{im:images_in_dataset}
\end{figure}

The dataset we use is the German Traffic Sign Recognition Benchmark  \cite{gtsrb_dataset,gtsrb_dataset_paper}. It contains more than 52.000 RGB images which are scaled to pixel size $32 \times 32$ in 43 different classes of German traffic signs. 75 percent of the data is used for training and validation and 25 percent for testing. One image of each of the 43 classes is shown in \autoref{im:images_in_dataset}

\section{Experiments}

In this section, we explain the implementation of the poisoning attacks and defences and discuss the chosen parameters. The full code is available online\footnote{ \url{https://github.com/lukasschulth/MA-Detection-of-Poisoning-Attacks}}.

\subsection{Poisoning Attacks}
We implement a targeted backdoor poisoning attack. We initially tried to attack the stop sign class since this would have the maximum adverse effect in practice. However, due to the relatively low amount of stop signs in the dataset and the needed change of labels, which would imply moving some stop signs to another class and therefore reducing the number of images in the stop sign class, we decided against this and instead chose to implement an attack for classifying speed limit 50 signs as speed limit 80 signs. From an attacker's perspective, this attack is somewhat weaker as its consequences in practice would be less catastrophic.

In \cite{badnets, AC} the used datasets also include a bounding box, which is used to determine the exact position of the pixels which are to be manipulated. Since bounding boxes are not available for our dataset, we inserted the pixel block containing the trigger at random in a window of $10 \times 15$ pixels in the center of the image.

Parameters for the attacks are the size of the backdoor trigger and the percentage of poisoned images in the poisoned class. We denote by $s$ the side length in pixels of the quadratic trigger pattern and consider $s\in \{1,2,3\}$ in our experiments. $p$ denotes the percentage of poisoned images in the attacked class.

\subsubsection{Standard Backdoor Poisoning Attacks}
For standard attacks, we fix the percentage $p$ and side length $s$ first. We then choose the number of speed limit 50 signs and change their label to speed limit 80 after adding a yellow-green pixel sticker such that the proportion in the speed limit 80 class is equal to $p$.

\subsubsection{Label-consistent Poisoning Attacks}
For label-consistent poisoning attacks, we first use a second network to reduce the performance of the targeted network when classifying the chosen images. We assume the attacker has full knowledge of the used network architecture and can use the same architecture to distort the images. Therefore, we used the same training setup and projective gradient descent with $n=5$ iterations and $step\_size=0.015$ to distort the images before adding the sticker.

\subsubsection{Label-consistent Poisoning Attacks with reduced sticker amplitudes}
For label-consistent poisoning attacks with reduced sticker amplitude, we fix the side length to $s=3$ and use the amplitude $amp$ as a new parameter, which is added or subtracted to the color channels depending on the pixel position inside the pixel block. We choose amplitudes of $amp\in\{256, 128, 64\}$.
Instead of placing the amplitude trigger of size $3\times 3$ pixels randomly, we ensure that its lower right pixel has a distance of 10 pixels to the right and bottom edge of the image. This is a modification of the sticker placement policy used in \cite{labelconsistent}. Placing the sticker in this way makes sure it is not cut off while using different data augmentation strategies (e.g. cropping) during training.

\subsection{Defences}
\noindent \textbf{Activation Clustering}\\
Before clustering the image data using $k$-means, we reduced the data to 10 dimension using PCA \cite{pca}.\\

\noindent \textbf{Layer-wise Relevance Propagation}\\
For the implementation of LRP, we used the work of \cite{moboehle} and adjusted it to the the inception modules. We also added propagation rules for BatchNorm-layers following the ideas of \cite{lapuschkin}. \\

\noindent \textbf{Calculation of Gromov Wasserstein Distances}\\ 
The entropy-regularized Gromov-Wasserstein distance is an approximation of the Gromov-Wasserstein distance that can easily be calculated by using the Python Optimal Transport Toolbox \cite{pot}. This library can also handle the calculation of GW-barycenters.

\section{Results}
In this section we compare the performance of Activation and Heatmap Clustering.

\begin{table*}
	\centering
	\begin{tabular}[h]{|c|c|ccc|ccc|}\hline

			 	&	&		& Detection Rate (AC)	& 		& 		&Detection Rate (HC) 	&		\\ 
			p 	& MASR 	& ACC 		& TPR 			& TNR 		& ACC 		& TPR 			& TNR 	\\ \hline
			
			33 	& 90.98	& 99.09 	& 97.24 		& 100.0  	&\textbf{99.58}	& 98.71			&100.0		\\
			15	& 83.15 & 96.42 	& 89.15			& 100.0 	&\textbf{98.91} & 96.69  		&100.0\\
			10	& 90.45 & 98.97 	& 89.7			& 100.0 	&\textbf{99.82} & 98.18			&100.00\\
			5	& 84.77 & 99.09 	& 84.15			& 99.87 	& \textbf{99.82} & 96.34		&100.00\\
			2	& 83.51 & \textbf{56.3}	& 100.00 		& 55.41 	& 55.33 	& 100.00		& 54.42\\ \hline
		\end{tabular}		
	\caption{Comparison of the performance of the detection algorithms in detecting label-consistent poisoning attacks for different percentages of poisoned data using the yellow-green pixel sticker with $s=3$.}
	\label{tab:Ergebnisse_CLPA}
\end{table*}

Before comparing both methods, we determine the performance for clustering on the raw data as a baseline. In this case, we try to detect standard attacks with quadratic stickers of side length $s=3$ and $33 \%$ manipulated data in the class in question. We use two different methods for dimensionality reduction (PCA\footnote{\url{https://scikit-learn.org/stable/modules/generated/sklearn.decomposition.PCA.html?highlight=pca\#sklearn.decomposition.PCA}} and FastICA\footnote{\url{https://scikit-learn.org/stable/modules/generated/sklearn.decomposition.FastICA.html}}). The results are shown in \autoref{tab:clustering_raw}.

\begin{table}[ht]
	\begin{center}
		\begin{tabular}{|c|c|c|} \hline
			& PCA & FastICA \\ \hline
			ACC	 & 	65.49 & 63.13 \\
			TPR		& 51.54 & 31.99 \\
			TNR	& 72.36 	&78.48\\ \hline 	 
		\end{tabular}
		\caption[Clustering result on raw data.]{Clustering result of clustering on raw image data reduced in dimensionality for standard attacks with 33 percent manipulated data and quadratic sticker of side length $s=3$.}	
		\label{tab:clustering_raw}
		
	\end{center}
\end{table}
We can see that in both cases the accuracy is about $2/3$. From the TPR we see that using PCA performs better than using FastICA, but still misses half of the poisoned data.
Regarding the TNR, we can see that in both cases about the same number of images are falsely classified as poisoned.

\subsection{Standard Backdoor Poisoning Attacks}
We  now compare Activation and Heatmap clustering for different standard attacks. In \autoref{tab:SPA_def_inv3_gwclustering} we can see the performance of both methods for attacks with a quadratic sticker of side length $s=3$ and percentages of manipulated data ranging from $33 \%$ down to $2 \%$.

Firstly, we can see that the attack success rate for all percentages of manipulated data is at $100 \%$. As we can see, both accuracy and true positive rate are around $99$ percent and above $90$ percent, respectively, for Heatmap Clustering and percentages of manipulated data above $2$ percent. Heatmap clustering has a close to perfect true negative rate, whereas the true negative rate for Activation Clustering is not that high. For $33\%, 15\%$ and $10\%$ manipulated data, the TPR of Activation clustering is also above 90 percent.

For both detection algorithms, none of the poisoned data points is correctly detected in the experiment with $2$ percent of the data poisoned. It is important to mention that the ASR is still at 100 percent.

\begin{figure*}
	\centering
	\includegraphics[height=0.2\textheight, width=\textwidth]{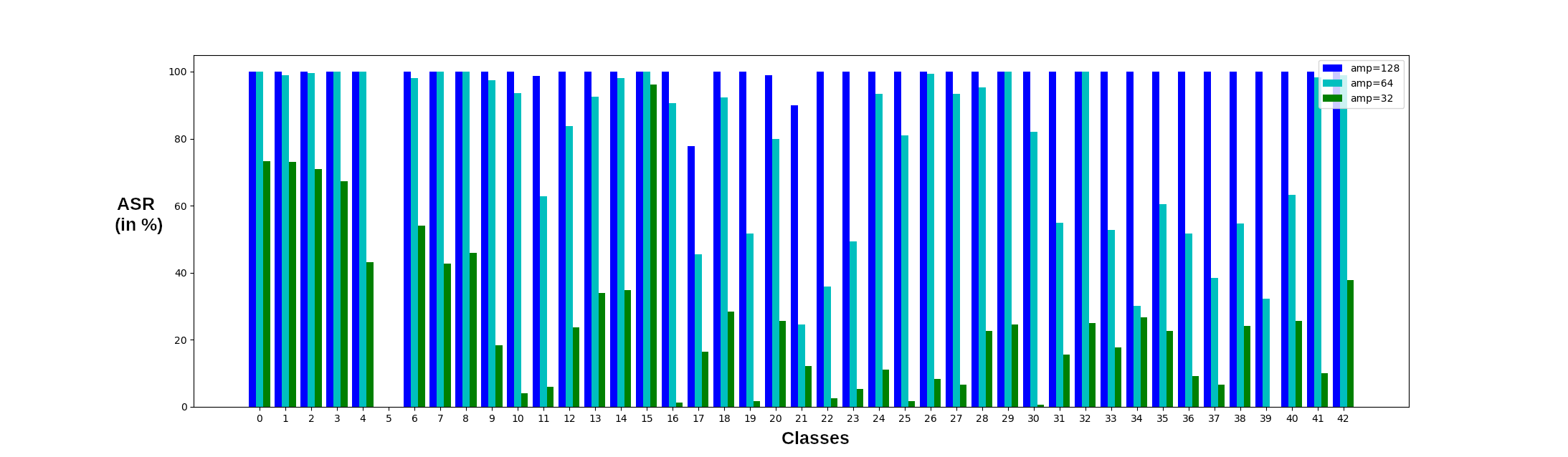}
	\caption{Attack Success Rate per class using an amplitude sticker and $33\%$ of poisoned data. We do not present an ASR for class 5 since this is the target class.}
	\label{fig:AER_proKlasse_CLPA_amp}
\end{figure*}

\begin{table*}
	
	\centering
	\begin{tabular}[h]{|c|c|ccc|ccc|} \hline
			&		&	& Detection Rate (AC)	& 		& 	&Detection Rate (HC)	&		\\
			Amplitude 	& MASR 	& ACC 			& TPR 		& TNR 	& ACC 			& TPR 		& TNR 	\\ \hline
			
			128 		& 99.18	& 99.45	 		& 98.35 	& 100.0 & \textbf{99.7}		& 99.08		& 100.0\\
			64		& 77.97 & \textbf{92.06} 	& 90.44		& 92.86 & 90.79 		& 74.08 	&99.01\\
			32		& 25.65 & 78.97 		& 74.26		& 81.28 & \textbf{81.58} 	& 58.46		& 92.95\\ \hline
	\end{tabular}
	
	\caption{Results of the detection of label-consistent poisoning attacks with triggers with reduced amplitude. $33\%$ of the data were poisoned.}
	\label{tab:Ergebnisse_CLPA_reduced}
\end{table*}

\subsection{Label-consistent Poisoning Attacks}
We now look at the detection rates in the case of label-consistent poisoning attacks.
Results are shown in \autoref{tab:Ergebnisse_CLPA}. We can see that both detection methods detect the attacks fairly well, where HC is missing a few poisoned data samples less, which results in higher TPR and slightly higher accuracy.

\subsection{Label-consistent Poisoning Attacks with reduced sticker amplitudes}
We now compare both detection algorithms for poisoning attacks with amplitude stickers with reduced amplitude.
Results for the detection of these attacks are shown in \autoref{tab:Ergebnisse_CLPA_reduced}.

We again look at the MASR. We can see that reducing the amplitude from $128$ to $32$ results in a strong decay of the MASR. In \autoref{fig:AER_proKlasse_CLPA_amp}, the attack rates for each individual class except the target class are shown for $33$ percent of poisoned data.

\subsection{Further experiments}
We noticed different results depending on the position of the trigger as well as the choice of images to manipulate. In order to compare both detection methods more reliably, we therefore ran the experiments in the case of label-consistent backdoor attacks for 10 times with different positions of the backdoor trigger and a random choice of the manipulated images and calculated mean and variance for the detection rates. We ran these experiments for $33\%$ and $15\%$ of poisoned data. The results are shown in \autoref{tab:gemittelt}.

From the results we see a clear advantage of Heatmap Clustering when considering mean accuracy, TPR and TNR. In particular, the standard deviation of the accuracy for Activation Clustering is nearly two orders of magnitude higher than for Heatmap Clustering. Most of this effect is due to the much higher standard deviation for the TNR. We conclude that our detection method based on heatmaps is more robust to the changes mentioned above than Activation Clustering.

\begin{table*}
	
	\centering
	\resizebox{\textwidth}{!}{

		\begin{tabular}{|l|c|ccc|ccc|}
			\hline
			 	&  		&	&	Detection Rate 		& 			& 			& Detection Rate 	& 		  \\
			 	&  		&	&	Activation Clustering 	& 			& 			& Heatmap Clustering	&		\\
			p	& 		& ACC 	& 	TPR 			& TNR  			& ACC 			& TPR 			& TNR 	 	\\\hline
			
			33.00	& $\mu$		& 90.36	& 94.41				& 87.16 		&99.82			& 99.46			& 100.00   \\ 
				& $\sigma ^2$	& 6.81 	& 2.98				& 9.96 			& 0.19			& 0.58 			& 0.00 \\ \hline
			15.00	& $\mu$		& 85.24	& 96.56				& 83.24 		& 99.89			& 99.32			& 99.98	   \\ 
				& $\sigma ^2$	& 10.75 & 1.98				& 12.88 		& 0.17			& 1.17 			& 0.024	\\ \hline
	\end{tabular}}
	
	\caption{Statistics for detection rates for both methods over 10 different runs. Attacks were performed using the standard sticker and $s=3$.}
	\label{tab:gemittelt}
\end{table*}

\section{Conclusion}
In this article, we adapted the idea given in \cite{unmaskingCH} for the detection of poisoning attacks on neural networks using a small version of an Inception network. Our experimental results show that HC performs better on the implemented attacks. The assumption of \cite{labelconsistent} that label-consistent poisoning attacks using an amplitude sticker as a trigger are impossible to detect was proven wrong for the neural network we considered.

In comparison to AC, HC has a drastically longer run time. This is a result of the computationally intensive calculation of the GWD. The Sinkhorn algorithm scales quadratically with the number of relevances chosen from two heatmaps.
Therefore, it might be conceivable to use AC as an initial check for possible poisoning attacks, and if the proportion of the resulting two clusters exceeds a certain threshold, one could use HC for further investigations. Besides, it might be possible to adapt the approach from \cite{CH} to reduce the run time of HC.

Another difficulty is the calculation of heatmaps using LRP in the first place. The different LRP backpropagation rules have to be implemented for each different type of layer in the used network.

A potential direction for follow-up work is to repeat the experiments for larger networks, for example a residual neural network (ResNet) \cite{he2016deep} as used in the experiments of \cite{labelconsistent}. Heatmap Clustering could also be tested on the attacks presented in \cite{toxic}, which introduces benchmarks for all known poisoning attacks.

{\small
\bibliographystyle{ieee_fullname}
\bibliography{heatmapclustering.bib}

\begin{thebibliography}{10}\itemsep=-1pt

\bibitem{bary_wasserstein_space}
Martial Agueh and Guillaume Carlier.
\newblock Barycenters in the wasserstein space.
\newblock {\em SIAM Journal on Mathematical Analysis}, 43(2):904--924, 2011.

\bibitem{imagenet_unhansed_v1}
Christopher~J Anders, Talmaj Marin{\v{c}}, David Neumann, Wojciech Samek,
  Klaus-Robert M{\"u}ller, and Sebastian Lapuschkin.
\newblock Analyzing imagenet with spectral relevance analysis: Towards imagenet
  un-hans' ed.
\newblock {\em arXiv preprint arXiv:1912.11425}, 2019.

\bibitem{CH}
Christopher~J Anders, Leander Weber, David Neumann, Wojciech Samek,
  Klaus-Robert M{\"u}ller, and Sebastian Lapuschkin.
\newblock Finding and removing clever hans: Using explanation methods to debug
  and improve deep models.
\newblock 2020.

\bibitem{imagenet_unhansed_v2}
Christopher~J. Anders, Leander Weber, David Neumann, Wojciech Samek,
  Klaus-Robert Müller, and Sebastian Lapuschkin.
\newblock Finding and removing clever hans: Using explanation methods to debug
  and improve deep models, 2019.

\bibitem{kmeans++}
David Arthur and Sergei Vassilvitskii.
\newblock k-means++: The advantages of careful seeding.
\newblock Technical report, Stanford, 2006.

\bibitem{LRP_first_paper}
Sebastian Bach, Alexander Binder, Gr{\'e}goire Montavon, Frederick Klauschen,
  Klaus-Robert M{\"u}ller, and Wojciech Samek.
\newblock On pixel-wise explanations for non-linear classifier decisions by
  layer-wise relevance propagation.
\newblock {\em PloS one}, 10(7):e0130140, 2015.

\bibitem{baehrens2010explain}
David Baehrens, Timon Schroeter, Stefan Harmeling, Motoaki Kawanabe, Katja
  Hansen, and Klaus-Robert M{\"u}ller.
\newblock How to explain individual classification decisions.
\newblock {\em The Journal of Machine Learning Research}, 11:1803--1831, 2010.

\bibitem{networkdis}
David Bau, Bolei Zhou, Aditya Khosla, Aude Oliva, and Antonio Torralba.
\newblock Network dissection: Quantifying interpretability of deep visual
  representations.
\newblock In {\em Proceedings of the IEEE conference on computer vision and
  pattern recognition}, pages 6541--6549, 2017.

\bibitem{biggio2012poisoning}
Battista Biggio, Blaine Nelson, and Pavel Laskov.
\newblock Poisoning attacks against support vector machines.
\newblock {\em arXiv preprint arXiv:1206.6389}, 2012.

\bibitem{LRP_DNN}
Alexander Binder, Sebastian Bach, Gregoire Montavon, Klaus-Robert M{\"u}ller,
  and Wojciech Samek.
\newblock Layer-wise relevance propagation for deep neural network
  architectures.
\newblock In Kuinam~J. Kim and Nikolai Joukov, editors, {\em Information
  Science and Applications (ICISA) 2016}, pages 913--922, Singapore, 2016.
  Springer Singapore.

\bibitem{moboehle}
Moritz Böhle, Fabian Eitel, Rodrigo~Bermúdez Schettino, and Leon Sixt.
\newblock Pytorch-lrp.
\newblock \url{https://github.com/moboehle/Pytorch-LRP}, 2019.

\bibitem{AC}
Bryant Chen, Wilka Carvalho, Nathalie Baracaldo, Heiko Ludwig, Benjamin
  Edwards, Taesung Lee, Ian Molloy, and Biplav Srivastava.
\newblock Detecting backdoor attacks on deep neural networks by activation
  clustering.
\newblock {\em arXiv preprint arXiv:1811.03728}, 2018.

\bibitem{evtimov2017robust}
Ivan Evtimov, Kevin Eykholt, Earlence Fernandes, Tadayoshi Kohno, Bo Li, Atul
  Prakash, Amir Rahmati, and Dawn Song.
\newblock Robust physical-world attacks on machine learning models.
\newblock {\em arXiv preprint arXiv:1707.08945}, 2(3):4, 2017.

\bibitem{pot}
R{\'e}mi Flamary, Nicolas Courty, Alexandre Gramfort, Mokhtar~Z. Alaya,
  Aur{\'e}lie Boisbunon, Stanislas Chambon, Laetitia Chapel, Adrien Corenflos,
  Kilian Fatras, Nemo Fournier, L{\'e}o Gautheron, Nathalie~T.H. Gayraud,
  Hicham Janati, Alain Rakotomamonjy, Ievgen Redko, Antoine Rolet, Antony
  Schutz, Vivien Seguy, Danica~J. Sutherland, Romain Tavenard, Alexander Tong,
  and Titouan Vayer.
\newblock Pot: Python optimal transport.
\newblock {\em Journal of Machine Learning Research}, 22(78):1--8, 2021.

\bibitem{goodfellow2014explaining}
Ian~J Goodfellow, Jonathon Shlens, and Christian Szegedy.
\newblock Explaining and harnessing adversarial examples.
\newblock {\em arXiv preprint arXiv:1412.6572}, 2014.

\bibitem{gromov2007metric}
Mikhail Gromov.
\newblock {\em Metric structures for Riemannian and non-Riemannian spaces}.
\newblock Springer Science \& Business Media, 2007.

\bibitem{badnets}
Tianyu Gu, Brendan Dolan-Gavitt, and Siddharth Garg.
\newblock Badnets: Identifying vulnerabilities in the machine learning model
  supply chain.
\newblock {\em arXiv preprint arXiv:1708.06733}, 2017.

\bibitem{he2016deep}
Kaiming He, Xiangyu Zhang, Shaoqing Ren, and Jian Sun.
\newblock Deep residual learning for image recognition.
\newblock In {\em Proceedings of the IEEE conference on computer vision and
  pattern recognition}, pages 770--778, 2016.

\bibitem{adam}
Diederik~P. Kingma and Jimmy Ba.
\newblock Adam: A method for stochastic optimization, 2014.

\bibitem{kistudie}
Dr.~Tom Kraus, Lene Ganschow, Marlene Eisenträger, and Dr.~Steffen Wischmann.
\newblock ErklÄrbare ki - anforderungen, anwendungsfälle und lösungen.
\newblock 2021.

\bibitem{kurakin2018adversarial}
Alexey Kurakin, Ian~J Goodfellow, and Samy Bengio.
\newblock Adversarial examples in the physical world.
\newblock In {\em Artificial intelligence safety and security}, pages 99--112.
  Chapman and Hall/CRC, 2018.

\bibitem{lapuschkin}
S. Lapuschkin.
\newblock Opening the machine learning black box with layer-wise relevance
  propagation.
\newblock 2019.

\bibitem{unmaskingCH}
Sebastian Lapuschkin, Stephan W{\"a}ldchen, Alexander Binder, Gr{\'e}goire
  Montavon, Wojciech Samek, and Klaus-Robert M{\"u}ller.
\newblock Unmasking clever hans predictors and assessing what machines really
  learn.
\newblock {\em Nature communications}, 10(1):1--8, 2019.

\bibitem{lipton}
Zachary~Chase Lipton.
\newblock The mythos of model interpretability. corr abs/1606.03490 (2016).
\newblock {\em arXiv preprint arXiv:1606.03490}, 2016.

\bibitem{madry2017towards}
Aleksander Madry, Aleksandar Makelov, Ludwig Schmidt, Dimitris Tsipras, and
  Adrian Vladu.
\newblock Towards deep learning models resistant to adversarial attacks.
\newblock {\em arXiv preprint arXiv:1706.06083}, 2017.

\bibitem{memoli2011gromov}
Facundo M{\'e}moli.
\newblock Gromov--wasserstein distances and the metric approach to object
  matching.
\newblock {\em Foundations of computational mathematics}, 11(4):417--487, 2011.

\bibitem{feature_vis}
Chris Olah, Alexander Mordvintsev, and Ludwig Schubert.
\newblock {Feature Visualization: How neural networks build up their
  understanding of images}.
\newblock \url{https://distill.pub/2017/feature-visualization/}, 2017.
\newblock [Online; Zugriff am 20.09.2021].

\bibitem{papernot2016transferability}
Nicolas Papernot, Patrick McDaniel, and Ian Goodfellow.
\newblock Transferability in machine learning: from phenomena to black-box
  attacks using adversarial samples.
\newblock {\em arXiv preprint arXiv:1605.07277}, 2016.

\bibitem{computationalOT}
Gabriel Peyre and Marco Cuturi.
\newblock Computational optimal transport.
\newblock {\em Foundations and Trends in Machine Learning}, 11(5-6):355--607,
  2019.

\bibitem{gwd_averaging_kernels}
Gabriel Peyr{\'e}, Marco Cuturi, and Justin Solomon.
\newblock Gromov-wasserstein averaging of kernel and distance matrices.
\newblock In {\em International Conference on Machine Learning}, pages
  2664--2672. PMLR, 2016.

\bibitem{lime}
Marco~Tulio Ribeiro, Sameer Singh, and Carlos Guestrin.
\newblock " why should i trust you?" explaining the predictions of any
  classifier.
\newblock In {\em Proceedings of the 22nd ACM SIGKDD international conference
  on knowledge discovery and data mining}, pages 1135--1144, 2016.

\bibitem{SHAP}
Marco~T{\'{u}}lio Ribeiro, Sameer Singh, and Carlos Guestrin.
\newblock "why should {I} trust you?": Explaining the predictions of any
  classifier.
\newblock {\em CoRR}, abs/1602.04938, 2016.

\bibitem{deeplearning_weather}
MG Schultz, Clara Betancourt, Bing Gong, Felix Kleinert, Michael Langguth, LH
  Leufen, Amirpasha Mozaffari, and Scarlet Stadtler.
\newblock Can deep learning beat numerical weather prediction?
\newblock {\em Philosophical Transactions of the Royal Society A},
  379(2194):20200097, 2021.

\bibitem{toxic}
Avi Schwarzschild, Micah Goldblum, Arjun Gupta, John~P Dickerson, and Tom
  Goldstein.
\newblock Just how toxic is data poisoning? a benchmark for backdoor and data
  poisoning attacks.
\newblock 2020.

\bibitem{gradcam}
Ramprasaath~R. Selvaraju, Michael Cogswell, Abhishek Das, Ramakrishna Vedantam,
  Devi Parikh, and Dhruv Batra.
\newblock Grad-cam: Visual explanations from deep networks via gradient-based
  localization.
\newblock {\em International Journal of Computer Vision}, 128(2):336–359, Oct
  2019.

\bibitem{shafahi2018poison}
Ali Shafahi, W~Ronny Huang, Mahyar Najibi, Octavian Suciu, Christoph Studer,
  Tudor Dumitras, and Tom Goldstein.
\newblock Poison frogs! targeted clean-label poisoning attacks on neural
  networks.
\newblock {\em Advances in neural information processing systems}, 31, 2018.

\bibitem{deeplift}
Avanti Shrikumar, Peyton Greenside, and Anshul Kundaje.
\newblock Learning important features through propagating activation
  differences.
\newblock In {\em International conference on machine learning}, pages
  3145--3153. PMLR, 2017.

\bibitem{gtsrb_dataset}
Johannes Stallkamp, Marc Schlipsing, Jan Salmen, and Christian Igel.
\newblock The {G}erman {T}raffic {S}ign {R}ecognition {B}enchmark: A
  multi-class classification competition.
\newblock In {\em IEEE International Joint Conference on Neural Networks},
  pages 1453--1460, 2011.

\bibitem{gtsrb_dataset_paper}
J. Stallkamp, M. Schlipsing, J. Salmen, and C. Igel.
\newblock Man vs. computer: Benchmarking machine learning algorithms for
  traffic sign recognition.
\newblock {\em Neural Networks}, (0):--, 2012.

\bibitem{integratedgradients}
Mukund Sundararajan, Ankur Taly, and Qiqi Yan.
\newblock Axiomatic attribution for deep networks.
\newblock In {\em International conference on machine learning}, pages
  3319--3328. PMLR, 2017.

\bibitem{goingdeeperwithconvolutions}
Christian Szegedy, Wei Liu, Yangqing Jia, Pierre Sermanet, Scott Reed, Dragomir
  Anguelov, Dumitru Erhan, Vincent Vanhoucke, and Andrew Rabinovich.
\newblock Going deeper with convolutions.
\newblock In {\em Proceedings of the IEEE conference on computer vision and
  pattern recognition}, pages 1--9, 2015.

\bibitem{spectral_signatures}
Brandon Tran, Jerry Li, and Aleksander Madry.
\newblock Spectral signatures in backdoor attacks.
\newblock {\em arXiv preprint arXiv:1811.00636}, 2018.

\bibitem{labelconsistent}
Alexander Turner, Dimitris Tsipras, and Aleksander Madry.
\newblock Label-consistent backdoor attacks.
\newblock 2019.

\bibitem{vayer2020contribution}
Titouan Vayer.
\newblock A contribution to optimal transport on incomparable spaces.
\newblock {\em arXiv preprint arXiv:2011.04447}, 2020.

\bibitem{pca}
Svante Wold, Kim Esbensen, and Paul Geladi.
\newblock Principal component analysis.
\newblock {\em Chemometrics and intelligent laboratory systems}, 2(1-3):37--52,
  1987.

\end{thebibliography}
}

\end{document}